\documentclass{article}

\usepackage[preprint, nonatbib]{neurips_2024}

\usepackage[utf8]{inputenc} 
\usepackage[T1]{fontenc}    
\usepackage{hyperref}       
\usepackage{url}            
\usepackage{booktabs}       
\usepackage{amsfonts}       
\usepackage{nicefrac}       
\usepackage{microtype}      
\usepackage{xcolor}         

\usepackage{graphicx}
\usepackage{algorithm} 
\usepackage{algpseudocode} 
\usepackage{amsmath}
\usepackage{xcolor}
\usepackage{setspace}
\usepackage{lineno}
\usepackage{subcaption}

\title{Adaptive Network Intervention for Complex Systems: A Hierarchical Graph Reinforcement Learning Approach}

\author{Qiliang Chen \\MAGICS lab \\ College of Engineering \\
Northeastern University \\
  Boston, MA 02115, USA \\
  \textit{chen.qil@northeastern.edu} \\\And
  Babak Heydari*\\MAGICS lab \\ 
  College of Engineering and Network Science Institute \\
  Northeastern University \\
  Boston, MA 02115, USA \\
  \textit{b.heydari@northeastern.edu} \\}

\begin{document}
\maketitle

\begin{abstract}

Effective governance and steering of behavior in complex multi-agent systems (MAS) are essential for managing system-wide outcomes, particularly in environments where interactions are structured by dynamic networks. In many applications, the goal is to promote pro-social behavior among agents, where network structure plays a pivotal role in shaping these interactions. This paper introduces a Hierarchical Graph Reinforcement Learning (HGRL) framework that governs such systems through targeted interventions in the network structure. Operating within the constraints of limited managerial authority, the HGRL framework demonstrates superior performance across a range of environmental conditions, outperforming established baseline methods. Our findings highlight the critical influence of agent-to-agent learning (social learning) on system behavior: under low social learning, the HGRL manager preserves cooperation, forming robust core-periphery networks dominated by cooperators. In contrast, high social learning accelerates defection, leading to sparser, chain-like networks. Additionally, the study underscores the importance of the system manager's authority level in preventing system-wide failures, such as agent rebellion or collapse, positioning HGRL as a powerful tool for dynamic network-based governance.

\end{abstract}



\onehalfspacing
\section{Introduction}
Governing complex multi-agent systems (MAS) has garnered increasing attention, particularly when network structures are incorporated, as they make these systems more realistic and relevant across various domains, such as socio-technical systems \cite{heydari2018guiding,sony2020industry, layton2016designing, o2017graph}, supply chain networks \cite{govindan2017supply, yazdandoost2024taxonomy, wu2023generative}, communication systems \cite{su2019resource, wu2016understanding, chaudhari2022co} and others \cite{lin2014network, panyam2019bio, dave2020designing}. Network structures not only dictate agent interactions—who engage with whom and how information is exchanged—but also play a pivotal role in shaping system-wide performance and driving the emergence of complex collective behaviors like coordination, cooperation, fairness, and trust \cite{trigeorgis1996real, chen2024sos, gianetto2013catalysts, mosleh2017fair}. These insights from network science open the door to a novel form of governance, where dynamically adjusting interaction patterns within the system can steer it toward desired outcomes. This approach complements more conventional methods of governance, such as classical engineering control or agent-level incentive designs from economics and social sciences, offering a powerful new tool for managing complex systems.

Despite the potential of network-based governance, several significant challenges complicate the effective management of multi-agent systems through network interventions. These challenges include (1) the complex and evolving dynamics of agent interactions, (2) the limited authority of the system manager, and (3) the vastness of the state and action spaces involved in decision-making.

The first challenge stems from the inherent complexity and evolving dynamics of these systems. Multi-agent systems typically involve interactions among individuals operating across varying levels of cooperation, coordination, and competition \cite{gianetto2015network, soria2018design, ji2022knowledge}. Each agent, acting autonomously, aims to optimize its own objectives within the environment, but these individual goals often conflict with the system’s collective welfare, leading to well-known social dilemmas \cite{poundstone1993prisoner, dawes1980social}. The dynamic nature of these systems further complicates matters, as network topologies are not static but evolve through both agent interactions and external interventions\cite{sha2014estimating}. Agents not only interact with their peers but also learn and adapt through social learning processes, observing and imitating the behaviors of others, particularly when those behaviors result in higher utilities \cite{bandura1977social}. As a result, these systems exhibit non-stationary dynamics, constantly changing and requiring the system manager to adopt highly flexible and adaptive governance strategies.

The second challenge arises from the limited authority of the system manager. In most real-world applications, a system manager’s ability to intervene is restricted by resource limitations and the level of compliance from the agents within the system, whether they are human or AI agents \cite{morato2014toward, mosleh2016resource, zhang2024early}. Excessive alterations to the network structure can incur prohibitively high resource costs, making it unfeasible for the system manager to continually change the network \cite{he2021people, van2022emergence}. Moreover, agents—particularly human ones—may resist directives that they perceive as overly controlling or misaligned with their own interests. As a result, the system manager must strike a careful balance between improving system performance and managing resources efficiently, all while operating under these constraints. Developing strategies that align with the limited authority of the manager becomes essential to navigating these challenges and effectively governing the system \cite{chen2022dynamic, mosleh2016distributed}.

The third challenge involves the vastness of both the state and action spaces in network interventions. As the number of agents $N$ in a network grows, the complexity of the network topology scales exponentially, with the number of potential configurations represented as $O(2^{N(N-1)/2})$. This makes it increasingly difficult for the system manager to monitor the system and make informed decisions about which links to add or remove \cite{trigeorgis1996real}. Additionally, the action space—the number of possible interventions available to the system manager—grows similarly in complexity, expressed as $O(N(N-1)/2)$. As a result, the system manager faces the daunting task of distilling critical information from an overwhelming number of possibilities, all within a limited number of training rounds. The combination of vast state and action spaces, along with the variability in agent features, creates a significant obstacle in learning efficient policies for network intervention.

\begin{figure*}[t]
   \centering
   \includegraphics[width=\textwidth]{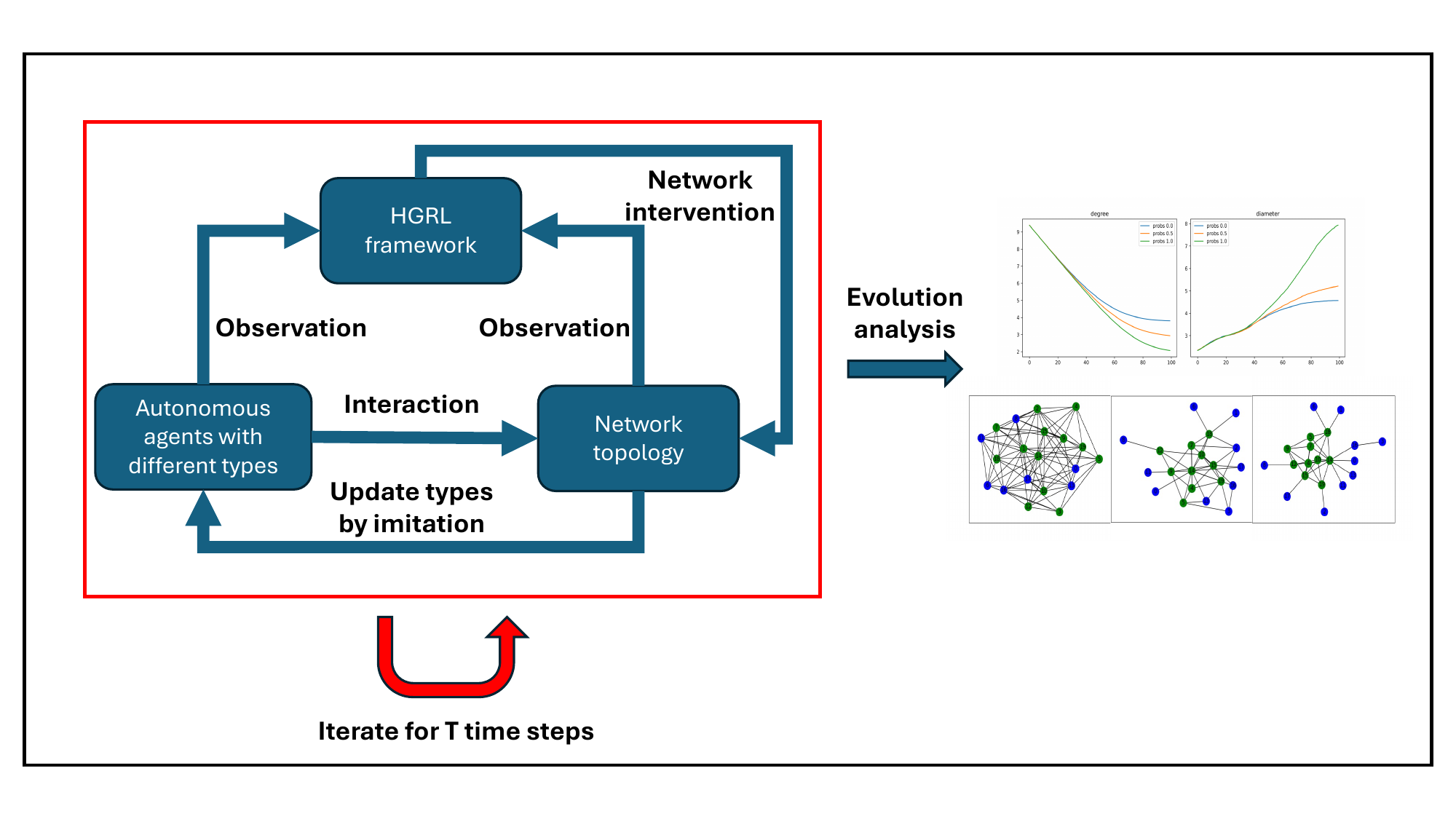}

   \caption{The diagram of the general process in the environment. We have 3 entities in the whole process: HGRL framework (details are shown in Figure \ref{fig:mesh1}), autonomous agents and network topology. In each round of the game, agents with different types interact with their first-order neighbors in the networks; the HGRL framework will observe the information from agents and network topology to make decision of network intervention by adding or deleting links in the network; agents will then imitate others with higher utilities under different imitation probabilities. This process will be iterated for several time steps and we can generate analysis of performance and network evolution from different perspectives.}
   \label{fig:general_diagram}
\end{figure*} 

To tackle these multifaceted challenges, we introduce a novel Hierarchical Graph Reinforcement Learning (HGRL) framework, which combines Graph Neural Networks (GNN), Reinforcement Learning (RL), and hierarchical structures. The HGRL framework allows the system manager to efficiently learn network intervention policies, even in the face of complex, evolving multi-agent systems with constrained authority. By utilizing a hierarchical approach, HGRL significantly reduces the complexity of both the state and action spaces, leading to more effective and scalable governance.

In the simulated environment we developed, the system manager adjusts the network structure using information from the entire system, while agents update their strategies through social learning, based on their observations and interactions with other agents. This iterative process allows the system to evolve over time, with the system manager aiming to optimize social welfare. Figure \ref{fig:general_diagram} provides a visual representation of this process.

Our evaluation shows that the HGRL framework consistently outperforms baseline models across a variety of conditions. Analysis of system behavior reveals that, with low levels of social learning, the HGRL manager preserves cooperation by forming a strong core-periphery network, where cooperators occupy the core. However, in high social-learning scenarios, defection spreads rapidly, resulting in a sparse, chain-like network structure. Our findings also highlight the importance of managerial authority in preventing system-wide failures, such as agent rebellion or collapse.

The paper is organized as follows: Section 2 reviews relevant literature, Section 3 provides a comprehensive introduction to the HGRL framework and our training methodology, Section 4 details the environment design and discusses both the performance results and the learned behaviors of the HGRL system manager, and Section 5 concludes the paper.

\section{Background}
This section reviews literature related to our paper. We begin by providing background information on network interventions in multi-agent systems, which is the central topic of this paper. Subsequently, we discuss papers that integrate Graph Neural Networks with Deep Reinforcement Learning, a methodology our approach also employs by combining these two methods.

\subsection{Network intervention in multi-agent system}
Comprehensive research in Network Science strongly supports the idea that the performance and behaviors of systems with network structures can be effectively managed through network intervention. Besides, numerous studies build on this premise, developing general models and methodologies applicable to various contexts.

Authors in \cite{li2018enabling} integrate graph and control theory to address challenges in structural controllability and optimal control of large-scale networks. Besides, Authors in \cite{ding2019key} develop a series of algorithms to identify key nodes in networks. Intervention in these nodes can significantly influence the overall system. Apart from cooperative scenarios, authors in \cite{zhang2018resilience} introduce a mathematical framework using signed graphs and dynamical system theories in multi-agent systems with 'coopetition networks', leveraging distributed interaction laws and neural network-based estimators for disturbance adaptation.

Additionally, several research efforts aim to solve network intervention tasks in specific areas, further expanding the scope and application of these methodologies. In the public health area, authors in \cite{siciliano2021strategies} propose a new framework for network interventions in public administration, drawing on Valente's 2012 topology and public health concepts, to enhance behavioral change and improve outcomes at social, organizational, and community levels. In the socio-technical systems, the study in \cite{ellis2020implementing} explores how networks and network formation influence the implementation of a self-management support intervention in a community setting.  In the infrastructure systems, the authors in \cite{hearnshaw2013complex} suggest the system manager build a supply chain with the presence of hub firms and redundancy and undertake a multi‐sourcing strategy or intermediation between hub firms to increase system resilience. 

Most existing studies in this field concentrate on specific domains, employing extensive domain knowledge to devise strategies for network intervention in multi-agent systems. However, the transferability of these strategies across various fields is limited. Our work is pioneering in addressing these common challenges we proposed in the context of network intervention problems in multi-agent systems with an innovative HGRL framework, designed to be versatile and applicable across a variety of application areas.

\subsection{Frameworks with combination of GNN and DRL}
The evolution of Reinforcement Learning (RL) has emerged as a universal solution for decision-making across various domains, yielding remarkable results in fields such as robotics \cite{schulman2017proximal,yu2020meta, poudel2020generative}, video gaming \cite{mnih2015human} and design and governance of engineering systems \cite{chen2021leveraging, chen2022dynamic}. Besides, Graph Neural Networks (GNNs) have demonstrated significant advancements in the perception of graph information in different fields like bio-medicine \cite{zhang2021graph}, computer vision \cite{cao2022applications}, and social science \cite{fan2019graph, gao2022hetinf, tomy2022estimating}. Consequently, when addressing decision-making in graph-based environments, the integration of GNNs with RL becomes a popular approach. 

Authors in \cite{almasan2022deep} proposed a GNN-based DRL agent to solve routing optimization problems and the results show that the DRL+GNN agent can outperform state-of-the-art solutions in the testing phase. To improve the network embedding task, authors in \cite{yan2020automatic} combined deep reinforcement learning with a novel neural network structure based on graph convolutional networks and proposed a new algorithm for automatic virtual network embedding. In chemical process design, authors in \cite{stops2023flowsheet} modeled chemical processes as graphs and used GNN and DRL to learn from process graphs, which is capable of designing viable flowsheets economically in an illustrative case study. In smart manufacturing, authors in \cite{yang2023combining} combined GNN and DRL to solve dynamic job shop scheduling problems, with higher effectiveness and feasibility than regular DRL. Authors in \cite{mckee2023scaffolding} tried to tackle the problem of promoting pro-social behaviors through network intervention with GNN plus DRL, and the results on human groups show the superiority of the framework. 

While the integration of Graph Neural Networks (GNNs) with Deep Reinforcement Learning (DRL) has been applied across various domains, its implementation in addressing network intervention issues remains underexplored. Our research marks a significant advancement in crafting effective network intervention strategies within multi-agent systems. Additionally, our proposed Hierarchical Graph Reinforcement Learning (HGRL) framework considerably simplifies the complexity of the state and action spaces, enhancing the efficiency of the learning process.

\section{Method}
\subsection{Partial Observable Markov Decision Process and Deep Q-Learning}
In our framework, we utilize a reinforcement learning approach based on the Partially Observable Markov Decision Process (POMDP) model, which is particularly well-suited for environments where the state space is not fully visible. A Partial Observable Markov Decision Process(POMDP) can be described using a tuple $<S, A, O, T, R>$. $S$ is a set of possible states. $A$ is a set of available actions. $O$ is observation perceived from the state of the system which may not fully cover all information of the state. $T$ is transition function, where $T$: $O \times A \rightarrow O$. $R$ is reward function, where $O \times A \rightarrow R$. The objective in POMDP is to learn the optimal policy $P_{\theta}$: $O \times A\rightarrow [0,1]$, which can maximize the expected return: $\sum_{t=0}^{T}\gamma^t{r}^t$ where $\gamma$ is a discount factor, $T$ is the time horizon and $r$ is the reward at each time step. 

Deep Q-learning, as outlined in \cite{mnih2015human}, is a widely utilized algorithm within the field of reinforcement learning (RL). In this approach, the agent first perceives an observation $o$, which is a representation of the environmental state $s$. Based on this observation, the agent utilizes the action-value function $Q^{\pi}(o, a) = E[R|o = o^t, a = a^t]$ under a policy $\pi$, and selects actions in a greedy manner to maximize the action value. The core objective of Q-learning is to refine the action value function towards the optimal policy. This is achieved by minimizing the loss function $L_{\theta} = E_{o,a,r,o^{\prime}}[(Q(o,a|\theta) - y)^2]$, where $y$ is defined as $r + \gamma \max_{a^{\prime}}Q^{\prime}(o^{\prime},a^{\prime})$. Here, $Q$ function is parameterized by $\theta$ using a Deep Neural Network. The observation at the next time step is denoted as $o^\prime$, and $a^\prime$ represents the action taken at that time. The target Q function, $Q^\prime$, which aids in stabilizing the learning dynamics, is periodically updated and is a clone of the operational Q function. 

\subsection{Graph neural network}
Graph Neural Network (GNN) \cite{zhou2020graph} is a class of deep learning models that operate on graphs and have shown promising results in various tasks such as node classification, link prediction, and graph classification. Besides, Graph Neural Networks (GNNs) have demonstrated significant potential for efficient learning on large graphs \cite{kipf2016semi}. Similar to Convolutional Neural Networks (CNNs), which utilize a filter that slides across an image, GNNs employ a neighborhood aggregation function. This function slides over the graph, thereby conserving computational resources when dealing with large graphs. GNNs also adopt a pooling strategy akin to CNNs, aggregating information for each node. Once the node embeddings are obtained, they can be utilized for various downstream tasks as required. The following paragraphs provide some details about GNNs.

A graph can be represented as $G = (V, E)$, where $V$ is a set of nodes and $E$ is a set of edges connecting these nodes. The goal of GNN is to learn a function $f(G, X)$, where $X$ is a matrix of node features, that maps the graph and its node features to a target variable. There are three main operations in GNN, which are message-passing, message aggregation, hidden representation update, and a readout operation, which converts all node embedding to satisfy specific requirements of tasks.

The message-passing function can be represented as:
$m_{v \to u}^{(l)} = M^{(l)}\left(h_u^{(l-1)}, h_v^{(l-1)}, e_{u,v}\right)$ where $h_u^{(l-1)}$ and $h_v^{(l-1)}$ are the hidden representations of nodes $u$ and $v$ at layer $l-1$, $e_{u,v}$ is the edge feature between nodes $u$ and $v$, $M^{(l)}$ is a message function that generates a message from node $v$ to node $u$ at layer $l$, and $m_{v \to u}^{(l)}$ is the message passed from node $v$ to node $u$ at layer $l$. The message-passing function can be modeled using deep neural networks. The Aggregation function can be written as: 
$a_u^{(l)} = \text{AGG}^{(l)}\left(\{m_{v \to u}^{(l)} | v \in \mathcal{N}(u)\}\right)$ where $\mathcal{N}(u)$ is the set of neighboring nodes of node $u$, $\text{AGG}^{(l)}$ is an aggregation function that aggregates the messages received by node $u$ at layer $l$. The update function can be represented as: $h_u^{(l)} = U^{(l)}\left(h_u^{(l-1)}, a_u^{(l)}\right)$ where $U^{(l)}$ is an update function that updates the hidden representation of node $u$ at layer $l$ based on the hidden representation $h_u^{(l-1)}$ from the last layer and the aggregated message $a_u^{(l)}$. The above equations are applied iteratively for multiple layers until the final representation of each node is obtained. The final node representations can then be used for various downstream tasks.

In our case, we need to represent the whole graph, which is similar to the operation in GNN-based graph classification, so the readout function can be written as $z_G = READOUT({h_v^{(l)} | v \in G})$ where $z_G$ is the graph-level representation obtained by aggregating the hidden representations of all nodes in the graph $G$ at the final layer $K$, and $READOUT$ is an aggregation function that combines the hidden representations of all nodes. There are various options for the $READOUT$ function, such as concatenation, max pooling, mean pooling, sum pooling and some LSTM-based pooling methods\cite{zhou2020graph}. We used the concatenation for the $READOUT$ function to aggregate the information of all nodes.

\begin{figure*}[t]
\includegraphics[scale=0.40]{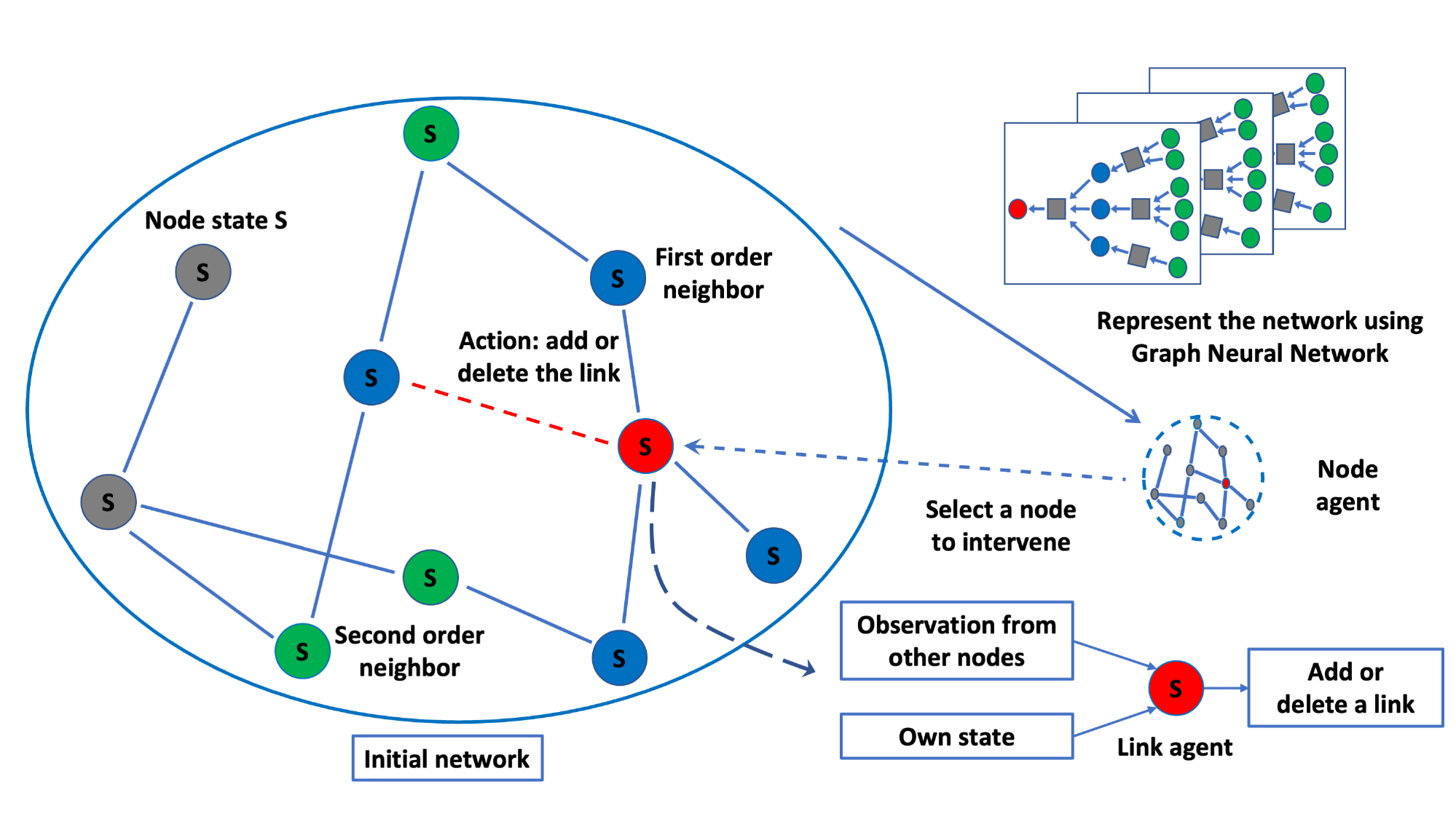}
\centering
\caption{General framework of HGRL. HGRL has two agents: node agent and link agent. The node agent will 
collect graph information using GNNs and select one node to intervene; the link agent will rely on the information related to the selected node to decide to add or delete links to this node. The action spaces for node agent and link agent are O(N), where N is the number of nodes in the network.}
\label{fig:mesh1}
\end{figure*}

\subsection{Hierachical Graph Reinforcement Learning framework}

Theoretically, to learn a policy of network intervention aimed at optimizing predefined goals, the standard Deep Q-Learning method (Flat-RL) can be employed, which utilizes system information to determine which links to add or remove in various situations. However, Flat-RL encounters significant challenges when applied to network structure interventions, primarily due to the rapidly increasing complexity of both the state and action spaces as the number of nodes increases. To illustrate, consider a network with N nodes, the potential structures for undirected networks scale as $O(2^{N(N-1)/2})$, which become intractable easily. Besides, the variability of potential node features will make it even more complex. The action space, involving the addition or deletion of links, has a complexity of $O(N(N-1))$. Consequently, both the state and action spaces in Flat-RL become exceedingly complex and potentially unmanageable as the network size increases.

To overcome this challenge, we propose the Hierarchical Graph Reinforcement Learning (HGRL) Framework. The overall process and structure of this framework are visualized in Figure \ref{fig:mesh1}. This framework leverages GNN for network representation, utilizes a hierarchical structure \cite{pateria2021hierarchical} and takes advantage of the separable cost idea in networks \cite{heydari2015efficient} to mitigate the issue of exploding state space and action space encountered in Flat-RL. Graph Neural Networks (GNN) efficiently embed network information, which reduces the state space complexity with less computational and memory demand. To tackle the action space reduction in link intervention, HGRL draws on the concept of separable costs in networks. Previous work used this idea by decomposing the link costs to the sum of costs from two separate end nodes, to find solutions for efficient network structures \cite{heydari2015efficient}. Inspired by this idea, links can be decomposed into pairs of end nodes. Instead of direct link selection, HGRL proposes selecting each end node independently. In undirected networks, the sequence of choosing these nodes is irrelevant. Thus, HGRL's approach involves a two-step hierarchical link intervention process. The initial step involves selecting one end node based on overall network information. The subsequent step depends on the first step, where the second end node is chosen based on the initially selected node and its associated information. This step effectively determines which links to add or remove from the chosen node. Essentially, HGRL simplifies the joint link selection process into two hierarchical, interdependent steps, significantly reducing the compounded action space from $O(N^2)$ to $O(N)$, while $N$ is the number of nodes in the network.

In the execution phase of the HGRL framework, the initial stage involves the node agent selecting the first end node based on graph information. This is achieved using a Graph Neural Network (GNN) to extract network data. Specifically, the node agent employs the message-passing process to embed each node as we introduced in section 3.2. This embedding considers both the hidden information of the node and its neighbors. The initial hidden features for layer 0 are the original node features. The embedded hidden information is then aggregated using mean pooling to obtain the first layer embedding. This embedding process can be repeated up to a total of $l$ layers, where $l$ is the number of GNN layers. The number of layers is critical as it defines the extent of the neighborhood perceived around each node; essentially, $l$ layers capture the information of the $l$-th order neighbors for each node. However, it is crucial to limit the number of layers because using too many layers can lead to an over-smoothing problem, where node representations become indistinguishable as the number of layers increases, as highlighted in the study by \cite{chen2020measuring}. For this application, a two-layer GNN is employed. After processing through these steps, the graph's embedding information, denoted as $h_{u_i}^{(l)} = GNN(u_i^{(0)}, E)$, is obtained. This information serves as the observation for the Q-function in the reinforcement learning algorithm. A fully connected neural network represents the Q-function, calculating the Q values for selecting each node: $Q_{node}((V,E), a_{u_i}) = Fullyconnect(h_{u_i}^{(l)})$, where GNN means the combination of the message passing, message aggregation, and hidden representation update we mentioned above. $u^{(0)}_i$ is the original node feature for node $i$, $a_{u_i}$ is selecting node $i$ to intervene and $h_{u_i}^{(l)}$ is the hidden information of node $i$ at $l$-layer GNNs. Then node agent will choose the node with the highest Q value. 

In the second stage of the HGRL framework, the link agent is tasked with selecting the second end node to either form a new link or delete an existing link with the node chosen in the first stage. The link agent's observation includes both the aggregated information of the neighboring nodes of the initially selected node and the existing connection status of that node selected from the first stage. The Q-function in this stage can be represented either using a Graph Neural Network (GNN) or a fully connected neural network. This flexibility is due to the reduced requirement for embedding the entire graph; instead, the focus is on calculating the Q values for all possible actions (adding or deleting links) from the node chosen in the first stage. Upon processing these calculations, the link agent will select the top $K$ actions that have the highest Q values ($K=1$ in our experiment). This selection is based on the intervention capability of the agent, as defined by the manager's authority. The process of conducting the second stage is similar to the first stage with different state space and action space accordingly. Importantly, the action spaces for both the node agent in the first stage and the link agent in the second stage are $O(N)$, where $N$ is the number of nodes in the network. This approach effectively mitigates the complexity of the action space, which in Flat Reinforcement Learning (Flat-RL) is $O(N^2)$.

The training process consists of two distinct phases. In the first phase, we train the link agent while employing a random strategy for the policy of the node agent. Initially, the node agent randomly selects one node. Subsequently, the link agent learns the policy to select a second node in order to determine which link to add or delete, based on the first selected node. Exploration is one of the important topics in RL; we use the $\epsilon$-greedy strategy to balance the exploration and exploitation in the training phase. The core idea is that when making decisions at each time step, the RL agent will have $epsilon$ probability to select actions randomly(exploration) instead of choosing the action with the highest Q values(exploitation), where $epsion$ will decrease gradually as the training proceed to make the policy converge. After making decisions at each time step, the trajectory of observation, actions, rewards, and next observations will be stored in a replay buffer. When updating the policy, a mini-batch of trajectory will be sampled from the replay buffer, and the Q-function will be updated to minimize the loss function, as we introduced in section 3.1. 

Once the link agent's training is complete, its policy will be fixed, and we will proceed to train the node agent. The general procedure is similar, but only the definitions of observation, action and reward for node agent are changed accordingly. The observation of the node agent will become all nodes' features and the connection status of the whole network. The action space of the node agent is selecting the first node strategically. The objectives for both the node and link agents are to improve social welfare in the system.

\section{Experiment and results}
In this section, we first introduce the environment we have designed in this work, which effectively captures the challenges of network intervention in the multi-agent systems we mentioned earlier. We then present and discuss the results of our proposed methods, examining both performance and behavioral aspects.

\subsection{Environment design}
To assess the effectiveness of the proposed HGRL framework, it's necessary to create an environment that encapsulates the key aspects of challenges as previously discussed. This environment should accurately represent the intricate interactions among agents within a multi-agent system structured as a network. Additionally, it must reflect the dynamic and complex nature of agent behavior and account for the constrained authority held by the system manager. First, we chose the Prisoner's Dilemma (PD) as the central mechanism in our network model, which is grounded in its proven effectiveness in representing the complexities of cooperative and competitive interactions \cite{axelrod1981evolution}. This fundamental game in Game theory exemplifies the conflict between individual rationality and collective benefit, making it an ideal framework for studying decision-making in socio-technical systems. This game has been widely studied in various disciplines, including economics \cite{arce2010economics}, sociology \cite{swedberg2001sociology}, and evolutionary biology \cite{le2007evolutionary}, to understand cooperative behavior among rational individuals. The nodes in the network are players, they will play PD with their neighbors using their own policies and receive rewards based on the utility matrix. 

Besides, agents can only observe their direct neighbors and imitate other's policies if the other's total utility is higher than theirs. The limited range of observation represents the partial observable challenges, and the process of imitation mirrors the social learning process, a fundamental aspect of human society, which is extensively discussed in Bandura's Social Learning Theory \cite{bandura1977social}. This imitation mechanism not only reflects the adaptability and evolution of strategies within a network but also underscores the importance of social influence and learning in shaping individual and collective behavior.

In addition, the environment also takes into account the limited authority of the system manager. This is achieved by constraining the system manager's ability to intervene in the network structure, allowing modifications to only a limited number of links at each time step, which, in our specific scenario, is restricted to one link. This aspect is crucial for ensuring the realism and applicability of our study to real-world socio-technical systems. In practical scenarios, system managers often do not possess infinite power to enforce rules or dictate behaviors within a network. Their influence is invariably circumscribed by legal, ethical, and practical constraints\cite{ostrom1990governing}. By considering the constraints on the system manager's authority, our model accurately represents the complexities and limitations that are inherent in the management and regulation of socio-technical systems. 

We provide some details of the environmental properties as follows. The agents will be randomly initialized with one of two policy types: "cooperator" or "defector". A cooperator consistently opts to cooperate, while a defector consistently chooses to defect during the game. In our environment, the utility matrix for the Prisoner's Dilemma (PD) is defined as follows: for mutual cooperation (CC), both players receive a utility of (-0.5,-0.5); if one cooperates and the other defects (CD), the cooperator receives -4 while the defector gets 0; in the reverse scenario (DC), the cooperator receives -4 and the defector 0; and for mutual defection (DD), both players receive a utility of (-3,-3). It's important to highlight that the HGRL model is adaptable to various utility matrices and capable of learning different policies accordingly. The utility matrix used here serves merely as an illustrative example, without any specific intent or preference for this particular set of utilities. The initial network structures are randomly established using the Erdős–Rényi model of network formation \cite{erdds1959random} with 0.5 probability of building links. The dynamics of agent interaction further involve a probability factor $p$. After interacting with their neighbors, each agent has a $p$ chance of evaluating one of their neighbors. This evaluation is based on comparing the total utilities gained in the last time step. If the neighbor's total utilities are higher, the agent will then imitate the type of that selected neighbor. The system manager can only add or delete one link at each time step because of limited authority, and its objective is to maximize the social welfare of the system.  We tested our framework on 10-node networks and 20-node networks. After network intervention, the disconnected networks were not allowed in our experiments.

\subsection{Experiment results -- performance}

\begin{figure*}[t]
   \centering
   \includegraphics[width=\textwidth]{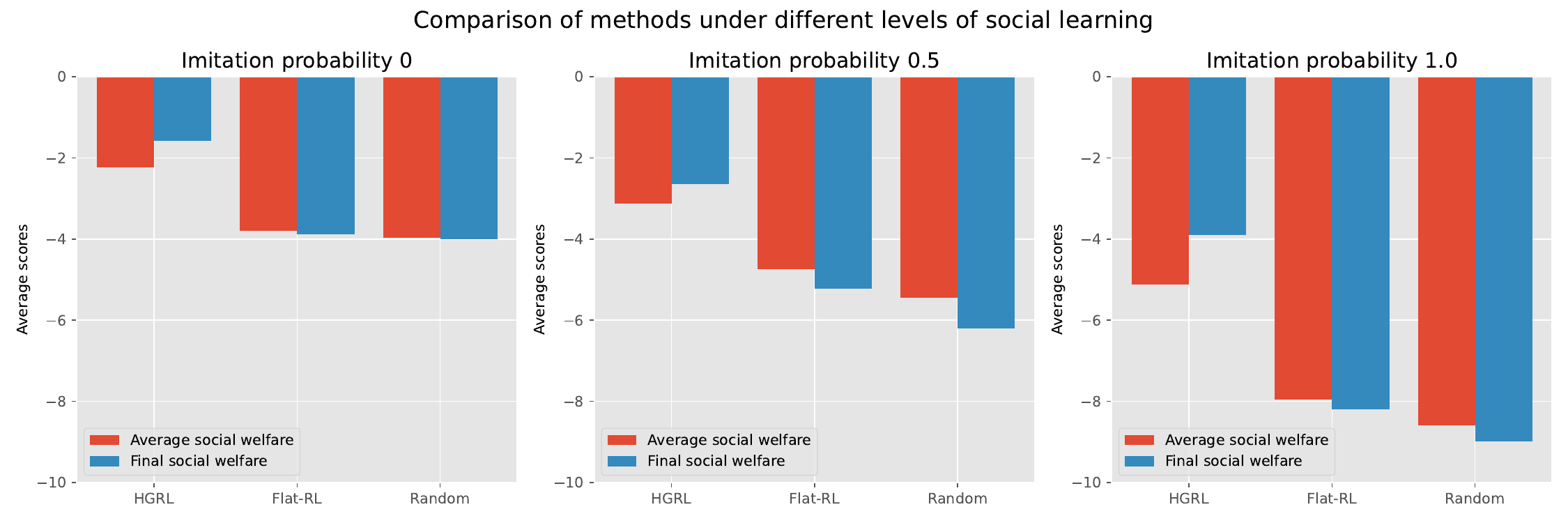}

   \caption{The performance comparison of HGRL, Flat-RL and random strategy in the system with 10 nodes. The figures show results from 3 scenarios with different probabilities of behavior imitation. Average social welfare (red bar) during the game and average final social welfare (blue bar) at the end of the game are used as metrics.}
   \label{fig:graphs_performance_10nodes}
\end{figure*}

\begin{figure*}[t]
   \centering
   \includegraphics[width=\textwidth]{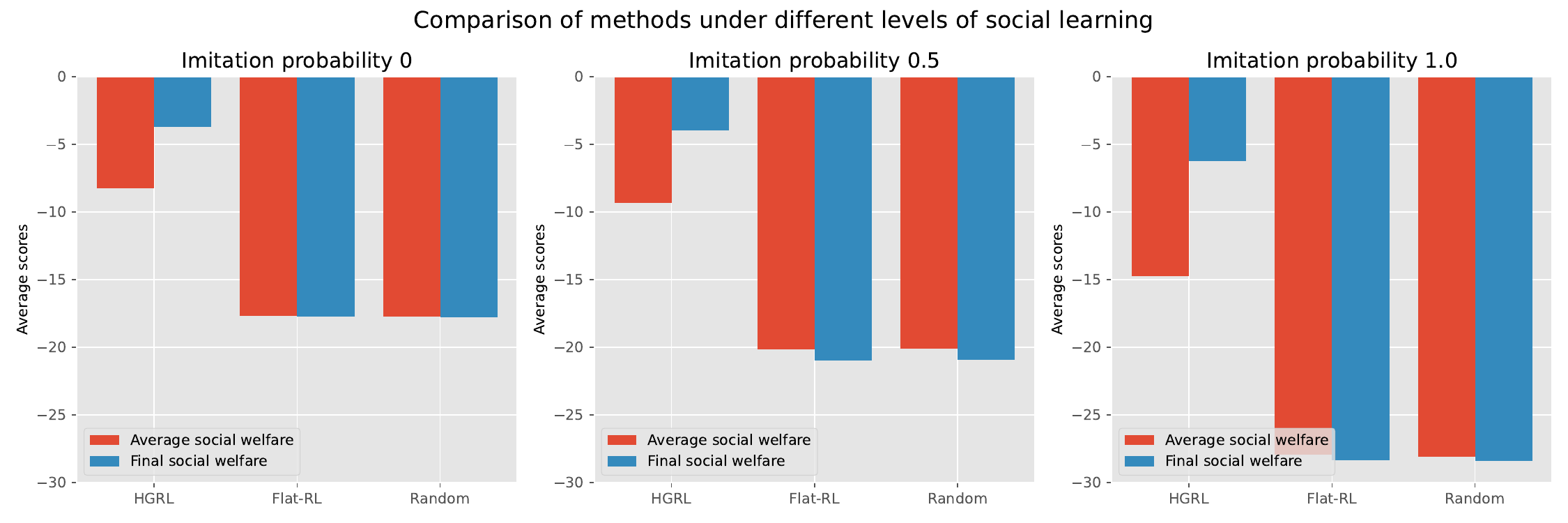}

   \caption{The performance comparison of HGRL, Flat-RL and random strategy in the system with 20 nodes. The figures show results from 3 scenarios with different probabilities of behavior imitation. Average social welfare (red bar) during the game and average final social welfare (blue bar) at the end of the game are used as metrics.}
   \label{fig:graphs_performance_20nodes}
\end{figure*}

In our study, we compared the performance of the HGRL model with that of Flat-RL and random strategy within the designed environment in both 10-node networks and 20-node networks. The HGRL model and Flat-RL model use similar sizes of parameters in neural networks. We trained the system manager for 10,000 rounds, with each round including 50 time steps for the 10-node networks and 100 time steps for the 20-node networks. The results presented are averaged from 1,000 test rounds. Our experiment was conducted under three different scenarios, each characterized by varying imitation probabilities $p$ of 0, 0.5, and 1. These probabilities represent different levels of imitation tendency among the agents. To assess the effectiveness of the models, we employed two metrics: average social welfare of the game and average final social welfare at the end of the game. The performance is also averaged by the number of agents in the system. Average social welfare offers a comprehensive evaluation of the models' effectiveness, encompassing not only the general performance but also the pace of improvement in social welfare, as a swifter enhancement results in a higher average social welfare. Meanwhile, the social welfare at the end of the game evaluates the models' ability to guide the system to an optimal or desirable state by the end of the game. This dual-metric approach allows for a comprehensive analysis of HGRL performance in managing the system with network structure. The results of the 10-node networks are in Figure \ref{fig:graphs_performance_10nodes}, and the results of 20-node networks are in Figure \ref{fig:graphs_performance_20nodes}. 

From the results, it's evident that under all scenarios, the HGRL method consistently outperforms both the Flat-RL and the random strategy in terms of both metrics. This superior performance of HGRL highlights its ability to learn more effective and adaptable policies for network structure intervention across various environmental settings. In contrast, Flat-RL can only surpass the random strategy with little improvement in small networks with 10 nodes, but in general, its performance is comparable to the random strategy, especially in large networks with 20 nodes. This suggests that with similar size of model parameters and training duration, Flat-RL struggles to develop a robust policy even in relatively small networks with just 10 nodes, which underscores the critical need and promising potential for HGRL in real-world applications. Given that many real-world multi-agent systems with large networks, Flat-RL's utility diminishes, whereas HGRL retains its capacity to learn effective policies by reducing the complexity of the state and action spaces. 

Besides, an insightful observation is that only the HGRL method demonstrates an ability to achieve higher social welfare at the end of games compared to the average social welfare. This observation reveals that HGRL uniquely facilitates a progressive enhancement in the system's social welfare throughout the game. In contrast, neither Flat-RL nor the random strategy exhibits this capability for gradual improvement. Various studies, such as those by \cite{gracia2012human, gianetto2015network, gianetto2016sparse, fulker2021spite}, have shown that defection can be 'contagious' in multi-agent systems due to the selfish behavior of each individual and also the modularity and cliques in network structures. Another key factor driving this trend is the bounded rationality of agents \cite{simon1957behavioral}, particularly humans, who are often swayed by immediate rewards at the expense of long-term benefits. While defection may cater to an agent's inherent preferences, offering higher immediate rewards, it can prompt other agents to defect rather than cooperate, ultimately undermining the overall social welfare of the system. This phenomenon underscores the importance of strategies like those learned in HGRL, which not only counter short-term defection tendencies but also enhance long-term cooperative behavior, thereby improving the overall health of the system.

Lastly, the performance of all methods decreases as the probability of imitation increases. Defection can be contagious and easily spread to the system because of social learning. When the probability of imitation increases, the speed of social learning becomes more rapid, and this tendency of defection spreading can easily dominate the whole system. In our experimental settings, if every agent becomes a defector and chooses to defect all the time, the social welfare will be the lowest, which justifies the results in the figure. In the real-world situation, we can also see some evidence. If every person is extremely selfish and only considers themselves, and when this behavior spreads rapidly, society will become unsympathetic and inconsiderate, the social welfare will decrease, and the system may even collapse.

\begin{figure*}[t]
   \centering
   \includegraphics[width=\textwidth]{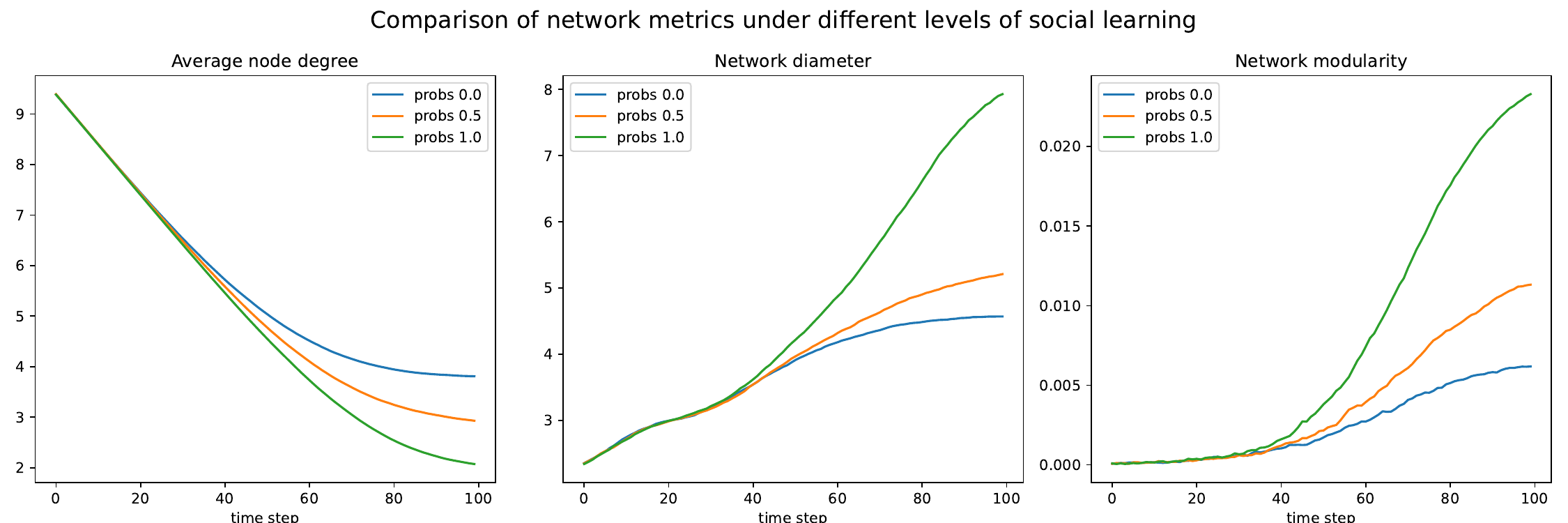}

   \caption{The evolution of different network metrics over time with various imitation probabilities. The figures show 3 network metrics: average node degree, network diameter, and network modularity. The x-axis is the time step.}
   \label{fig:network_metrics}
\end{figure*} 

\subsection{Experiment results -- behavior}
In the previous section, we demonstrated the superior performance of the proposed HGRL model over Flat-RL in various aspects, then it is equally crucial to examine the learned behavior of the HGRL manager. As AI models evolve, concerns about the "black-box" problem become increasingly pronounced. Understanding the decision-making process of the HGRL model could provide valuable insights into the development of more efficient and transparent AI governance strategies.

We only analyzed the behavior of the HGRL manager and the evolution of systems in 20-node networks. Three metrics were used to study the evolution of the networks over time under varying imitation probabilities: average node degrees, network diameter, and network modularity. The results are presented in Figure \ref{fig:network_metrics}. Generally, the curves for different imitation probabilities begin to diverge after 30 time steps. This pattern indicates that the influence of varying levels of social learning becomes significant after 30 time steps. Consequently, the systems initially exhibit similar trends but gradually diverge under the flexible intervention of HGRL's policies. 

Specifically, the average node degrees decrease through time in all scenarios, with higher social learning scenarios resulting in lower average node degrees. The initial random networks form several irrational connections, such as Cooperator-Defector and Defector-Defector links. Deleting these links benefits all scenarios, leading to generally sparser networks. In scenarios with higher social learning, it is more difficult for the system to conserve cooperators. As a result, these systems tend to be sparser which can reduce connections between defectors. Additionally, the network diameter generally increases over time, with scenarios of higher social learning resulting in larger network diameters. This pattern indicates that the networks tend to become more chain-like over time. Specifically, in scenarios with an imitation probability of 1.0, the networks conclude with 2 node degrees and a network diameter of 8 on average, essentially forming chain-like networks with few branches. Finally, the network modularity in all scenarios tends to increase over time, with higher social learning scenarios resulting in higher modularity. This indicates that agents in the system become more separated during evolution. In scenarios with lower imitation probability, more cooperators can be conserved and the presence of more connections to cooperators is beneficial to social welfare, which tends to form a large single community with lower network modularity in these scenarios.

\begin{figure*}[h!]
   \centering
   \begin{subfigure}[b]{0.3\textwidth}
      \includegraphics[width=\textwidth]{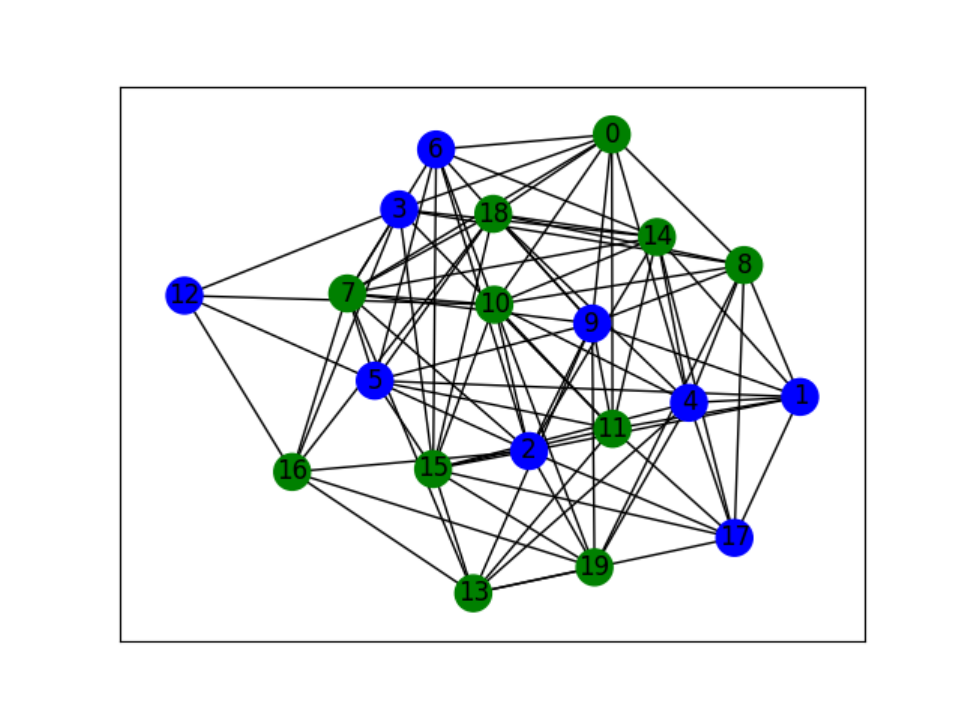}
      \caption{Initial state with 0.0 imitation probability}
      \label{fig:graph1}
   \end{subfigure}
   \hfill
   \begin{subfigure}[b]{0.3\textwidth}
      \includegraphics[width=\textwidth]{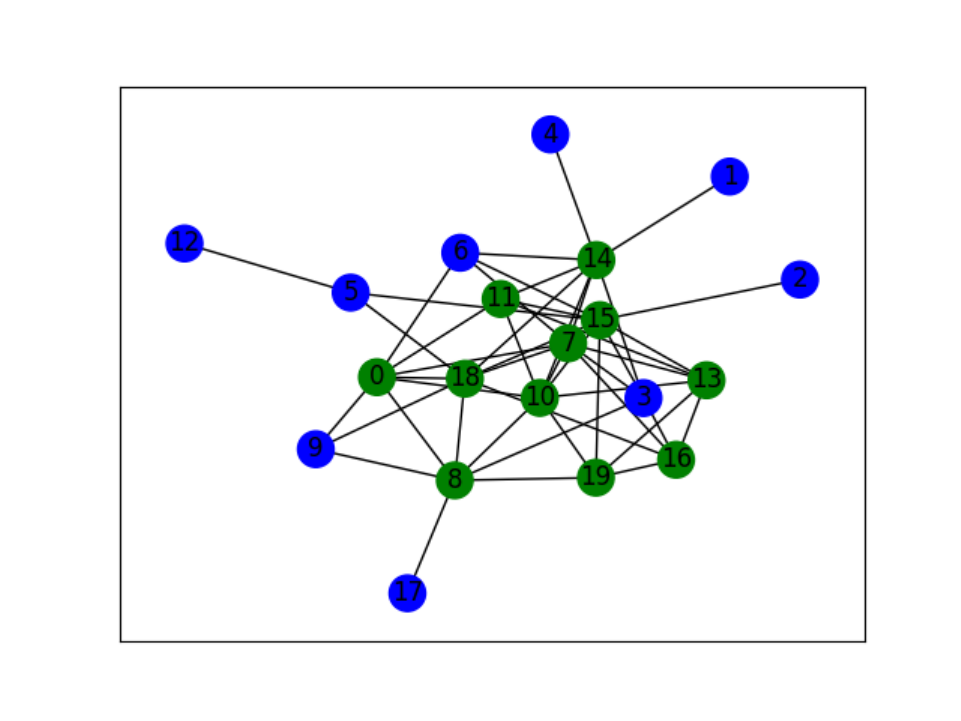}
      \caption{Middle state with 0.0 imitation probability}
      \label{fig:graph2}
   \end{subfigure}
   \hfill
   \begin{subfigure}[b]{0.3\textwidth}
      \includegraphics[width=\textwidth]{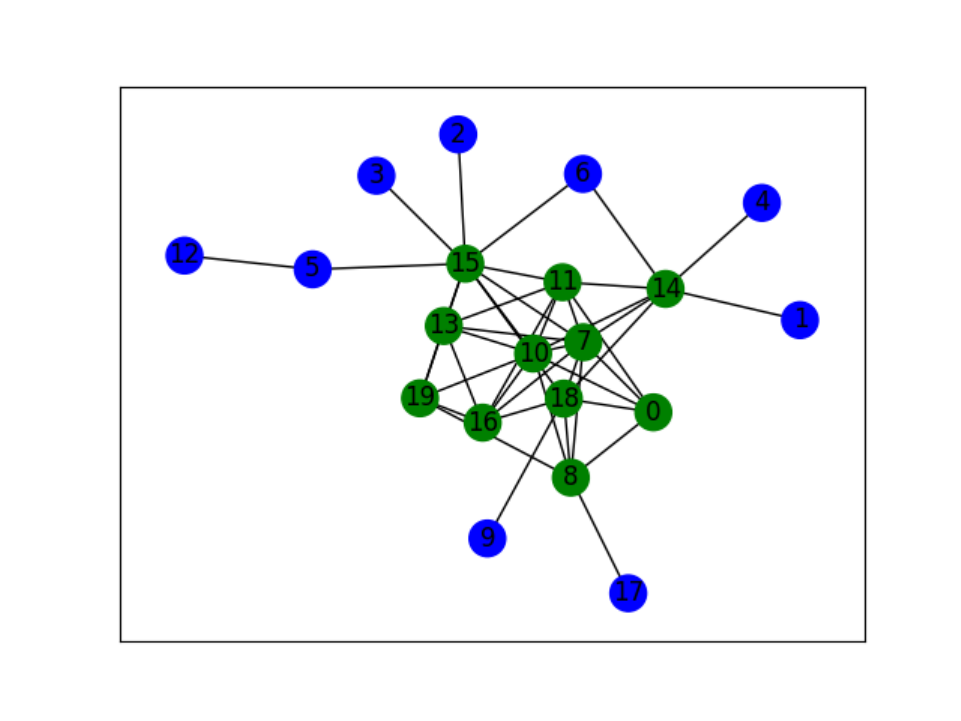}
      \caption{Final state with 0.0 imitation probability}
      \label{fig:graph3}
   \end{subfigure}

   \begin{subfigure}[b]{0.3\textwidth}
      \includegraphics[width=\textwidth]{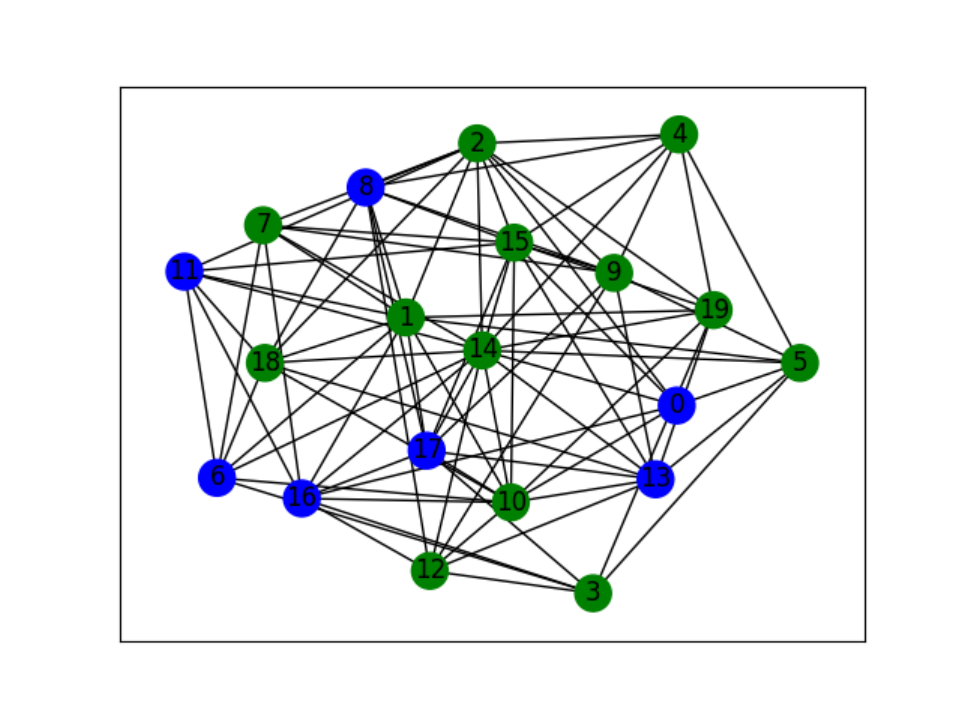}
      \caption{Initial state with 0.5 imitation probability}
      \label{fig:graph4}
   \end{subfigure}
   \hfill
   \begin{subfigure}[b]{0.3\textwidth}
      \includegraphics[width=\textwidth]{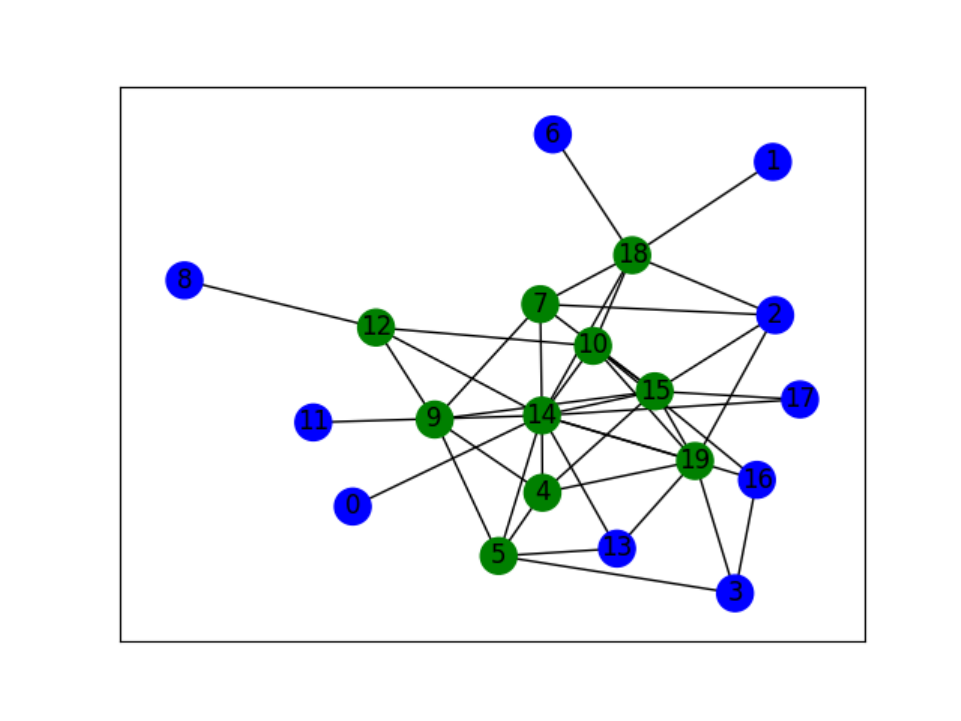}
      \caption{Middle state with 0.5 imitation probability}
      \label{fig:graph5}
   \end{subfigure}
   \hfill
   \begin{subfigure}[b]{0.3\textwidth}
      \includegraphics[width=\textwidth]{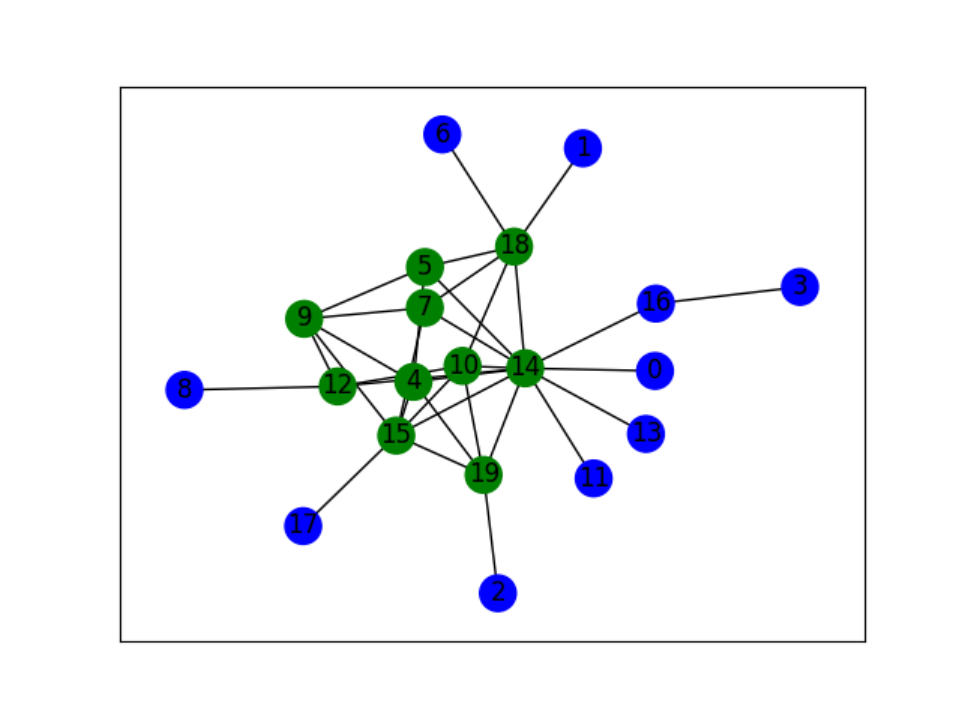}
      \caption{Final state with 0.5 imitation probability}
      \label{fig:graph6}
   \end{subfigure}

   \begin{subfigure}[b]{0.3\textwidth}
      \includegraphics[width=\textwidth]{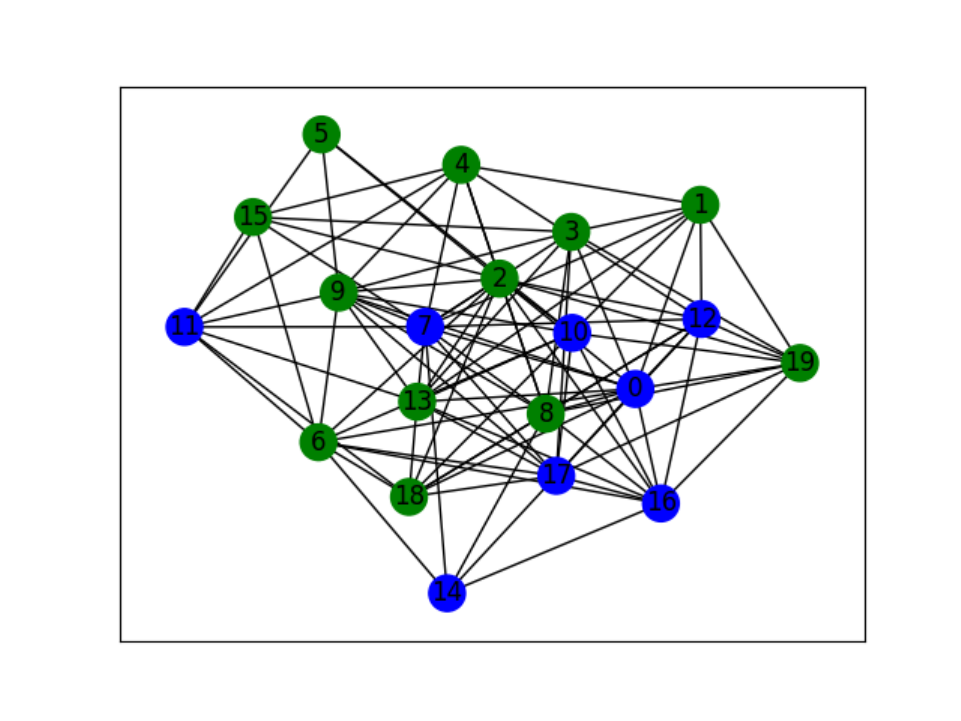}
      \caption{Initial state with 1.0 imitation probability}
      \label{fig:graph7}
   \end{subfigure}
   \hfill
   \begin{subfigure}[b]{0.3\textwidth}
      \includegraphics[width=\textwidth]{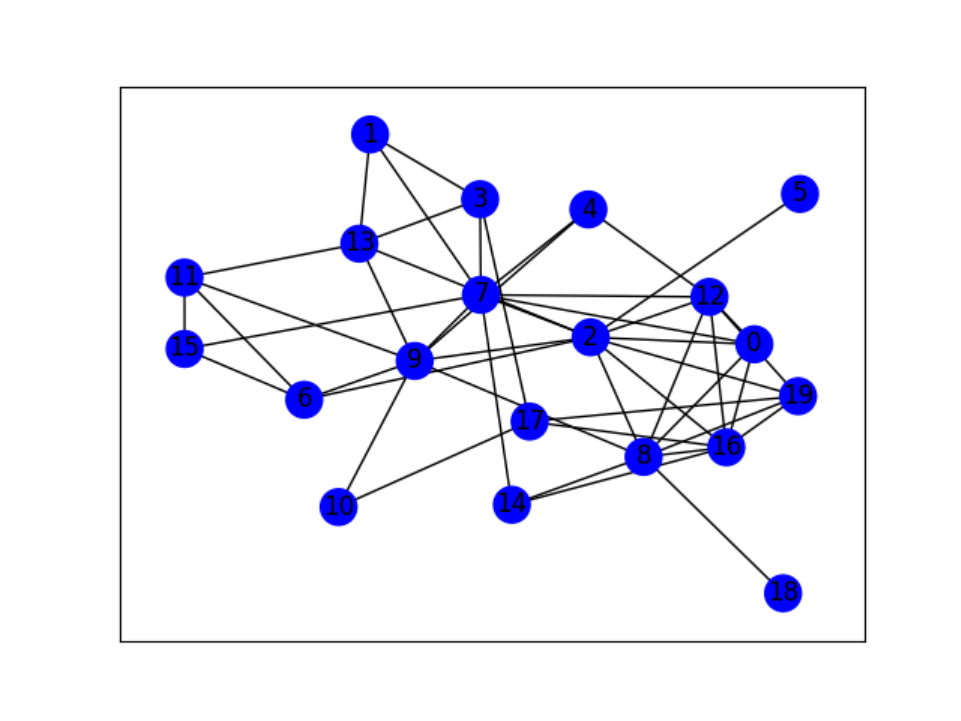}
      \caption{Middle state with 1.0 imitation probability}
      \label{fig:graph8}
   \end{subfigure}
   \hfill
   \begin{subfigure}[b]{0.3\textwidth}
      \includegraphics[width=\textwidth]{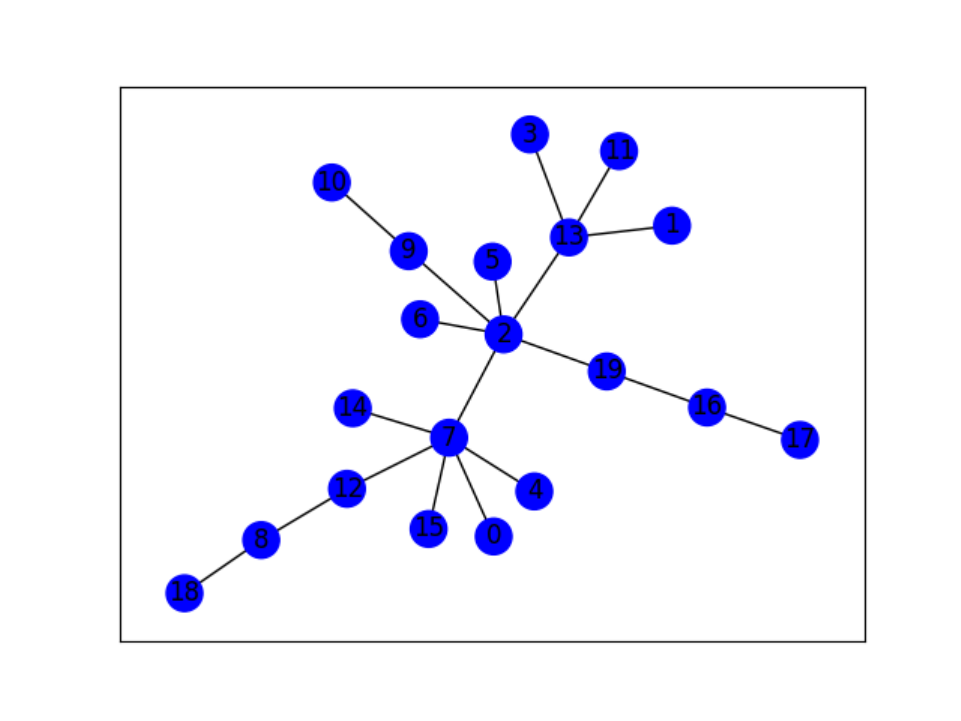}
      \caption{Final state with 1.0 imitation probability}
      \label{fig:graph9}
   \end{subfigure}

   \caption{The snapshots of the system evolution with 20 nodes in the initial, middle and final phases of the game under 0, 0.5 and 1 imitation probabilities. The green agents are cooperators, and the blue agents are the defectors.}
   \label{fig:3x3graphs}
\end{figure*}

We also visualized some snapshots of how the network evolves through time, we randomly selected one random seed as an example to take a look at the learned behaviors in 20-node scenarios. We took some snapshots of the system evolution under three different scenarios with imitation probability similar to previous settings, which can be seen in Figure \ref{fig:3x3graphs}. The first row illustrates the system's evolution under a scenario with 0 imitation probability, where agent types remain constant. Observing snapshots from different times during the game, it becomes apparent that the HGRL manager has learned to strategically manipulate network connections. This strategy involves fostering connections among cooperators, while simultaneously deleting links both between defectors and between cooperators and defectors. As a result, the final state of the system becomes a core-periphery network: a densely connected group of cooperators as cores and more loosely connected defectors. It's important to note that our environmental settings prohibit the disconnection of the network. Since interactions between cooperators and defectors, as well as between defectors themselves, tend to negatively impact social welfare, the framework will prevent these situations by disconnecting those interactions and promoting the existence of cooperator-cooperator connections.

The second row showcases the system's evolution in a scenario with a 0.5 imitation probability, where agents gradually modify their types through social learning. During this process, defection swiftly spreads among the agents, with 3 cooperators out of 13 shifting to defectors in mid-game. In contrast to the 0 imitation probability case, where disconnecting between cooperators-defectors and deleting connections between defectors are similarly prioritized, the HGRL manager in this scenario focuses more on breaking links between cooperators and defectors. This approach is understandable given that defection, with its temptation of higher immediate rewards, spreads via interactions between cooperators and defectors. While deleting connections among defectors does enhance social welfare in the short term, the cooperators are at risk of being "contaminated" by defection through social learning. This shift can lead to a long-term decline in social welfare, as defection becomes more dominant and agents increasingly opt to defect. The HGRL manager, therefore, adopts a forward-looking strategy that prioritizes long-term welfare over immediate gains. 

Interestingly, despite the spread of defection, it does not completely overpower the system by the game's end. There remain 10 cooperators due to the HGRL manager's strategic interventions. The final state reveals a robust cooperator community, resilient to defection despite some interactions with defectors. This cooperator community maintains a higher overall social welfare, safeguarding its members from deviating towards defection. To counter the spread of defection in this dynamic social learning environment, the HGRL manager's strategy is twofold: first, it targets critical links between cooperators and defectors for disconnection, and second, it rapidly builds a strong cooperator community before they become influenced by defectors. This dual-aspect approach not only preserves more cooperators but also elevates the system's overall social welfare.

The third row provides insights into scenarios where the imitation probability is set to 1.0, indicating a highly aggressive adaptation of types through social learning. In such a scenario, the transition from cooperators to defectors happens rapidly. By mid-game, all cooperators have already converted to defectors. Given these circumstances, the HGRL manager’s only viable strategy is to delete connections between defectors, aiming to mitigate the deteriorating situation. Consequently, at the game's end, we observe a sparse chain-like network comprising solely defectors. In this scenario, the authority of the system manager is deliberately constrained, allowing only minor alterations — one link change per time step. Given the rapid spread of defection, this limited authority proves insufficient to manage the situation effectively. Despite the manager's efforts to disconnect defectors, which may marginally preserve social welfare, defection ultimately prevails, leading to the system's collapse.

This outcome offers valuable insights, particularly in the context of real-world applications. It underscores the importance of appropriately balancing the level of managerial authority. Excessive authority can negatively impact agent behavior, especially when agents are human. In such cases, agents might refuse to follow commands, feeling coerced into actions against their will. Conversely, insufficient authority risks leading to an anarchic system where negative outcomes can be inevitable in some scenarios, and all agents suffer. This delicate balance is crucial in ensuring effective governance and maintaining a stable, functional multi-agent system.

\section{Conclusion}
This paper introduces the Hierarchical Graph Reinforcement Learning (HGRL) framework, designed to govern multi-agent systems with network structures through strategic network interventions. This framework is adaptable to a variety of initial network configurations and agent characteristics, accommodating the dynamics of social learning. Importantly, it operates under a realistic premise of limited authority, making it applicable to real-world systems where full control is often impractical. A key innovation of the HGRL framework is its ability to address the exponential complexity inherent in the state and action spaces of network structure interventions. It employs Graph Neural Networks (GNN) to manage the vast state space effectively and reduces the $O(N^2)$ action space of traditional Flat Reinforcement Learning (Flat-RL) to $O(N)$, due to its hierarchical structure.

Furthermore, the paper introduces the creation of an environment that encapsulates critical aspects of multi-agent systems with network structures. This environment not only simulates strategic interactions among agents but also incorporates the element of social learning and constrained managerial authority. From the results in the testing phase, the HGRL framework demonstrates superior performance over conventional Flat-RL in enhancing overall system performance and in guiding systems to states of higher social welfare. The improvement is consistently observed across various scenarios, each characterized by differing levels of imitation behavior in the social learning process. The learned behaviors of the HGRL framework are analyzed, affirming the rationality of its learned strategies. These strategies generate valuable heuristics that can be applied more broadly. Although the HGRL manager cannot prevent system collapse in cases of intense social learning, this outcome highlights the importance of authority levels in real-world applications and strategical governance of multi-agent systems with the complex dynamics of agents and their interactions.

One potential future work involves manipulating the information flow as a tool for network intervention. This approach aims to not only conserve the initial cooperators but also convert defectors to cooperators under non-trivial situations. Manipulating information as an intervention tool is advantageous because it is subtler and less noticeable to users, facilitating easier implementation. In the current framework, the policy of network structure intervention is trained to optimize social welfare, and the social learning process is influenced by the change of network structure dependently as indirect effects. Adding the idea of information manipulation, we can separate the social learning process and train another manager who recommends which agents to observe and imitate with reasonable constraints. Additionally, the imitation probability can be seen as the transparency or trustworthiness of the information. With a higher imitation probability, the information that agents can receive is more trustworthy, and they will base on it to update their strategy with less doubt. It is promising to consider changing the imitation probability dynamically to see how it can promote social welfare.

\bibliographystyle{unsrt}

\bibliography{arxiv.bib} 

\end{document}